\documentclass{article}
\usepackage{arxiv}
\usepackage{microtype}
\usepackage{hyperref}
\usepackage{url}
\usepackage{booktabs}
\usepackage{amsmath}
\usepackage{nicefrac}
\usepackage{xcolor}
\usepackage{natbib}
\usepackage{graphicx} 
\usepackage{hyperref}
\usepackage{cleveref}
\usepackage{tcolorbox} 

\usepackage{enumitem}
\usepackage{amsmath}

\newcommand\blfootnote[1]{%
  \begingroup
  \renewcommand\thefootnote{}\footnote{#1}%
  \addtocounter{footnote}{-1}%
  \endgroup
}

\definecolor{cornflowerblue}{rgb}{0.39, 0.58, 0.93}
\hypersetup{
    colorlinks=true,
    linkcolor=cornflowerblue,
    filecolor=magenta,      
    urlcolor=teal,
    citecolor=cornflowerblue,
    pdftitle={Benchmarking ChatGPT on Algorithmic Reasoning},
    pdfpagemode=FullScreen,
    }

\title{Benchmarking ChatGPT on Algorithmic Reasoning}

\author{Sean McLeish \\
\emph{University of Maryland}\\
\texttt{smcleish@umd.edu}
\And 
Avi Schwarzschild \\
\emph{Carnegie Mellon University}\\
\texttt{schwarzschild@cmu.edu}
\And 
Tom Goldstein\\
\emph{University of Maryland}\\
\texttt{tomg@cs.umd.edu}
}

\colmfinalcopy 
\begin{document}

\maketitle

\begin{abstract}
We evaluate ChatGPT's ability to solve algorithm problems from the CLRS benchmark suite that is designed for GNNs.
The benchmark requires the use of a specified classical algorithm to solve a given problem.
We find that ChatGPT outperforms specialist GNN models, using Python to successfully solve these problems.
This raises new points in the discussion about learning algorithms with neural networks and how we think about what out of distribution testing looks like with web scale training data.\blfootnote{Code available at 
\href{https://github.com/mcleish7/CLRS4LM}{\texttt{github.com/mcleish7/CLRS4LM}}
.}
\end{abstract}

\section{Introduction}
A number of recent works propose methods for neural algorithm synthesis. 
For algorithm learning on general graph data structures, DeepMind offers the CLRS benchmark comparing different algorithm synthesis agents for sorting, searching, dynamic programming, path-finding, and more. 
Many domains across computer vision and natural language processing have recently found that large, generalist foundation models can out-compete specialized systems on common benchmarks, and we observe that algorithm synthesis is no different. 
We show that GPT-4, when provided with a code interpreter, can solve our procedurally generated language version of CLRS benchmark to a level beyond currently published (GNN-based) solutions.

Recent work proposes several methods for neural algorithm synthesis using convolutional networks, GNNs and transformers, we highlight the ones most related to our work.
\citet{schwarzschild2021can},  \citet{bansal2022endtoend}, and  \citet{mcleish2023re} focus on end-to-end learning from data alone, using convolutional networks for reasoning tasks such as solving mazes.
There is also a large body of work \citep{mankowitz2023faster, dudzik2022graph, ibarz2022generalistlearner, bevilacqua2023neural, jain2023neural, rodionov2023neural, jayalath2023recursive, mirjanic2023latent, jung2023triplet, bohde2024markov, numeroso2023dual, georgiev2023neural, minder2023salsa} that consists of GNN models designed to solve algorithmic reasoning tasks, primarily focusing on the CLRS benchmark \citep{clrs-velickovic}.
Some particular examples include works that increase generalisation of reasoning within GNNs by employing techniques such as: recursion \citep{jayalath2023recursive}, using a looped transformer \citep{de2024simulation} or leveraging the duality in these problems \citep{numeroso2023dual}.

The thirty benchmark tasks in the CLRS suite are designed primarily for GNN pipelines \citep{clrs-velickovic} and are used most notably to benchmark GNNs \citep[e.g.][]{ibarz2022generalistlearner}.
Rather than use GNNs, we provide GPT-4 with each specific benchmark problem in natural language and a minimal description of the desired algorithmic goal, for example the algorithm name or a basic outline of the steps.
We find that when tasked with solving these algorithmic problems and asked to use a particular algorithm, ChatGPT can often write and execute the appropriate code in Python.
Executing code enhances the reasoning abilities of language models \citep[e.g.][]{gao2023pal,yang2024if}, especially over long trajectories.

\section{Benchmark Performance}

We focus on the tasks in the CLRS benchmark suite.
\citet{clrs-velickovic} select thirty algorithms from the CLRS textbook \citep{CLRSBOOK}, which are commonly taught in undergraduate algorithms courses and include examples like Prim's and Kruskal's Minimum Spanning Tree algorithms. 
\citet{clrs-velickovic} provide both training and testing data for each problem comprising input-output pairs as well as hints for each sample in the dataset.
The hints, corresponding to intermediate steps in the algorithm, are designed to allow for teacher forcing during training.
We do not use these hints nor provide them in our adaptation of the dataset.
In this report, we present comparisons where we test ChatGPT on samples from the testing and training splits.
(More details on which samples we use are available in the documentation in our code repository.) 
A full list of the thirty algorithms can be found in the original benchmark paper \citep{clrs-velickovic}, as well as in the labels of the figures below.
The designers of this benchmark actually suggest this dataset is of use for sequence to sequence models.
They write,
``while we format the data in a way that clearly
favours graph neural network executors, it can be easily
adapted for different types of neural architectures; for example, sequence to sequence models \citep{NIPS2014_seq2seq},'' \citep{clrs-velickovic}. 
We adapt their data in just this way for ChatGPT.

Our experiments are easy to describe, but require a lot of data wrangling to execute and our dataset is now available for further exploration into how LLMs can handle these tasks.
In short---we pose each of the CLRS problems as a word problem, including nested lists where arrays need to be described, and ask ChatGPT to execute a particular algorithm to solve the problem, either providing the algorithm name or a very minimal description where there may be confusion around the specific variation of the algorithm to be used.
In Example Prompt 1, we show one example of a prompt used in testing ChatGPT on Bubble Sort problems.
Note that this problem is phrased as simply as possible and provides the model with very little structure, but some natural language description of what to do.

\begin{tcolorbox}[title=\textbf{Example Prompt 1:} Bubble Sort Algorithm, colback=gray!5, colframe=gray!75, rounded corners]
{\footnotesize \textbf{System Prompt:}\\ You are a helpful assistant for solving and explaining classical coding problems.\\
\textbf{Context:} 
Perform Bubble Sort on this list [0.72322, 0.6891, 0.54337, 0.53711, 0.80969, 0.79958, 0.84777, 0.19036, 0.20027, 0.77366, 0.56553, 0.2689, 0.47936, 0.67466, 0.68423, 0.82139] and output the order of the indices, starting with initial indices [0, 1, 2, 3, 4, 5, 6, 7, 8, 9, 10, 11, 12, 13, 14, 15]. I cannot run code. You should show as much work as possible, at least the first step, and run until the sorting process is complete. The last line of your output should be the solution to the problem. If this is from running code, you should restate the output in our conversation.
}
\end{tcolorbox}

\begin{figure}[t!]
    \begin{center}
    \includegraphics[width=\textwidth]{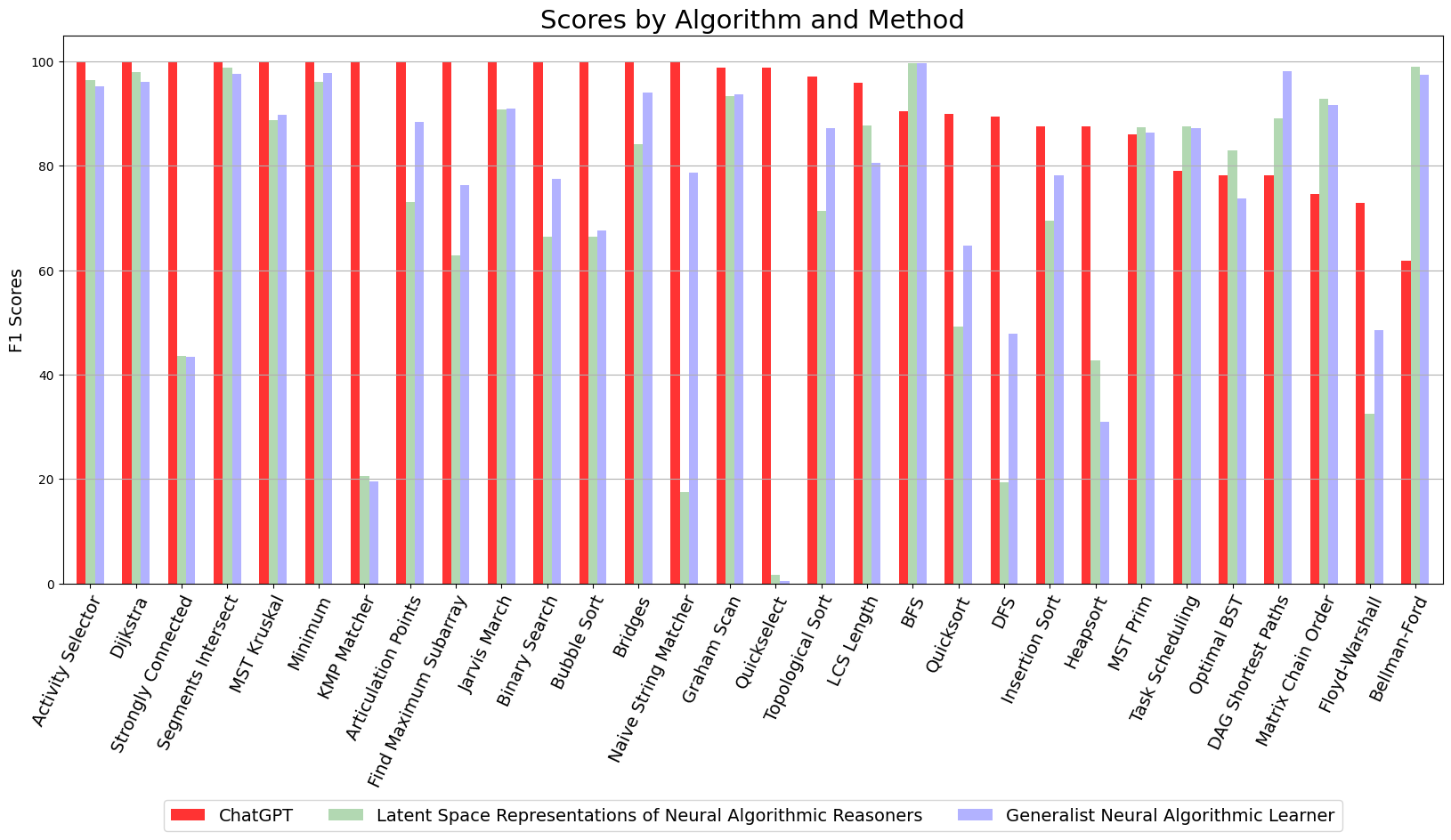}
    \end{center}
    \caption{
    Comparison to results presented by \citet{mirjanic2023latent} and \citet{ibarz2022generalistlearner} on testing data from CLRS. Note that ChatGPT is better on more than two thirds of the tasks and always roughly competitive with state-of-the-art GNN methods. We exclude examples where ChatGPT returns a question or there is an error in the OpenAI system.
    }
    \label{fig:test-comp}
\end{figure}

\paragraph{The Details} These tasks are split into eight general categories: sorting, searching, divide and conquer, greedy, dynamic programs, graphs, string matching, and geometry.\footnote{See the CLRS benchmark documentation for more detail at \href{https://github.com/google-deepmind/clrs}{\texttt{github.com/google-deepmind/clrs}}.}
We do not encode the CLRS problem inputs into natural language in any particularly creative way, instead we give ChatGPT actual arrays written out.
For example, for the string based problems, like Naive String Matcher and Knuth–Morris–Pratt String Matcher,
the inputs are drawn from an alphabet of size four and are one-hot encoded.
We input these one-hot encodings directly into ChatGPT, as shown in Example Prompt 2.
However, some of the problems have outputs that are encoded specifically for GNNs and hard to phrase in natural language. 
We preprocess these to transform them into the input-output pairs we need for LLM testing.  
For example, the format of the Longest Common Subsequence outputs in CLRS designed for GNNs to output one-hot encoded feature vectors at every node, but to make the requested output from ChatGPT easier to parse, we collapse those representations to integer indices and request a simpler two-dimensional array (more detail in \Cref{subsec:lcs}).

For some problems, we do not explicitly request the solution to the problem but a part of the solving process in order to determine if the model is using the required algorithm.
For example, for the Matrix Chain Order problem whose final output is an ordering of multiplications, we request an intermediate product of the algorithm called the split matrix as output.

\begin{figure}[t!]
    \centering
    \includegraphics[width=\textwidth]{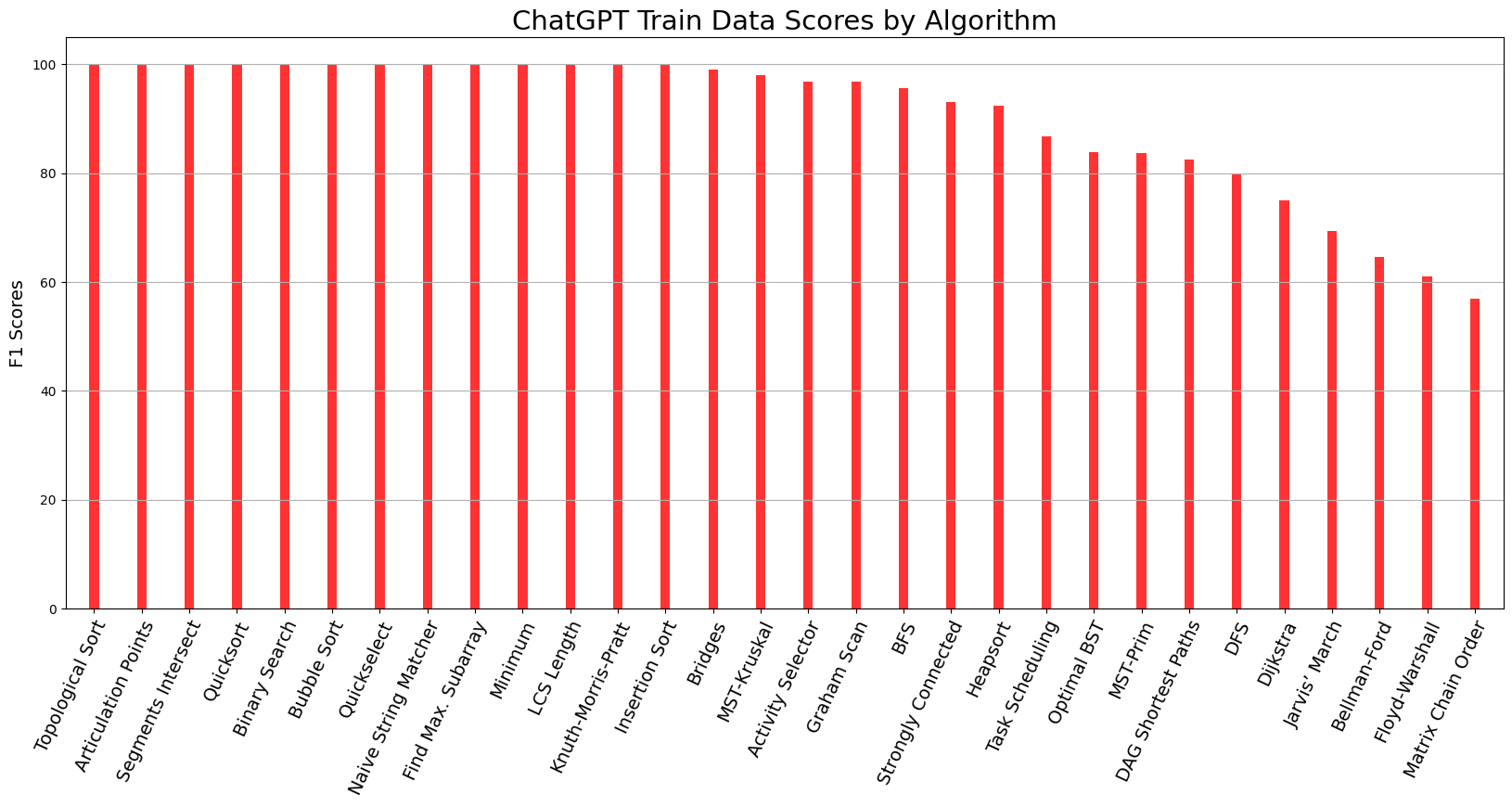}
    \caption{
    Individual F1 scores for training data from CLRS which comprises. We exclude examples where ChatGPT returns a question or an error from the OpenAI API.
    }
    \label{fig:train-F1}
\end{figure}

In \Cref{fig:test-comp}, we show performance on the testing split from CLRS. 
These examples are larger problems, for example the lists to sort are longer and the graphs to traverse have more nodes.
We compare ChatGPT to two existing GNN methods.
The first method is proposed by \citet{mirjanic2023latent} and we use the best of the models listed in Table 3 in their paper. 
The second is from \citet{ibarz2022generalistlearner} and we compare to their Triplet-GMPNN results from Table 2 in their paper.
In \Cref{fig:train-F1}, we present the F1 scores for ChatGPT on the training split, i.e. smaller size problems.
Note that \Cref{fig:train-F1} does not include results from the GNN methods as they do not share their training accuracies.

\begin{tcolorbox}[title=\textbf{Example Prompt 2}: One-Hot Encoding Style Problems, colback=gray!5, colframe=gray!75, rounded corners]
{\footnotesize \textbf{System Prompt:}\\ You are a helpful assistant for solving and explaining classical coding problems.\\
\textbf{Context:} 
Perform the Knuth-Morris-Pratt string matching algorithm on [[0, 1, 0, 0], [0, 0, 1, 0], [1, 0, 0, 0], [0, 0, 1, 0], [0, 1, 0, 0], [0, 0, 1, 0], [0, 0, 0, 1], [1, 0, 0, 0], [0, 1, 0, 0], [1, 0, 0, 0], [1, 0, 0, 0], [0, 1, 0, 0], [0, 1, 0, 0], [0, 0, 0, 1], [1, 0, 0, 0], [0, 0, 0, 1]] and [[0, 0, 1, 0], [0, 1, 0, 0], [0, 0, 1, 0], [0, 0, 0, 1]], where the characters of the string are one hot encoded from a size 4 vocabulary. Return the beginning index at which the strings overlap. If you write python code, the first code block should only be you defining the strings. I cannot run code. You should show as much work as possible, at least the first step, and run until the process is complete. The last line of your output should be the solution to the problem. If this is from running code, you should restate the output in our conversation.
}
\end{tcolorbox}

We conduct all experiments with single prompts and no feedback and find that on 24 of the problems ChatGPT performed better than the specially trained GNN approaches. 
See \Cref{sec:example-prompts} for additional examples of prompts from the training split of the CLRS benchmark; test sample prompts are structured identically, but contain larger problems (bigger arrays and longer lists).
We also include a sample ChatGPT output for Quick Sort in \Cref{sec:example-prompts}.
On some of the problems where ChatGPT performs worse, the outputs expected in the CLRS benchmark encode a lot of information about decision making within the algorithms that may not apply outside of the benchmark.
In particular, there are occasionally multiple correct ways to execute an algorithm---like which branch is traversed first in BFS.
Although the general rule for the CLRS benchmark is to choose the lowest index, giving this instruction in the ChatGPT context window can lead to a false interpretation of other parts of the algorithm.
For example, it may also enforce this rule on the queue used in BFS leading to an incorrect decision elsewhere in the algorithm.
However, in keeping with the strict nature of the benchmark we use the solutions as they are listed as ground truth and mark any other trajectories incorrect.

For some problems in the suite, the algorithm in question is not the most direct or efficient choice in general.
In theses cases, ChatGPT often tries to optimise its output, meaning it implements a more optimal solution than the given algorithm and thus is marked incorrect.
This is especially present in the task scheduling problem where we ask the model to execute the specific algorithm proposed by \citet{lawler1985travelling} but ChatGPT solves the scheduling problem but with another approach.

\begin{figure}[t!]
\centering
    \includegraphics[width=.49\textwidth]{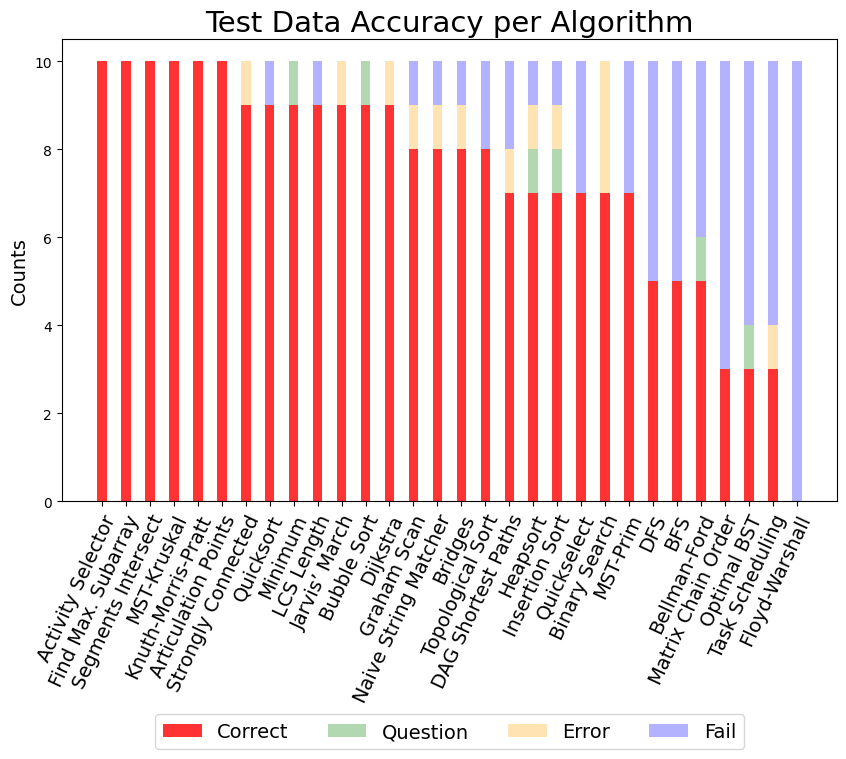}\hfill
    \includegraphics[width=.49\textwidth]{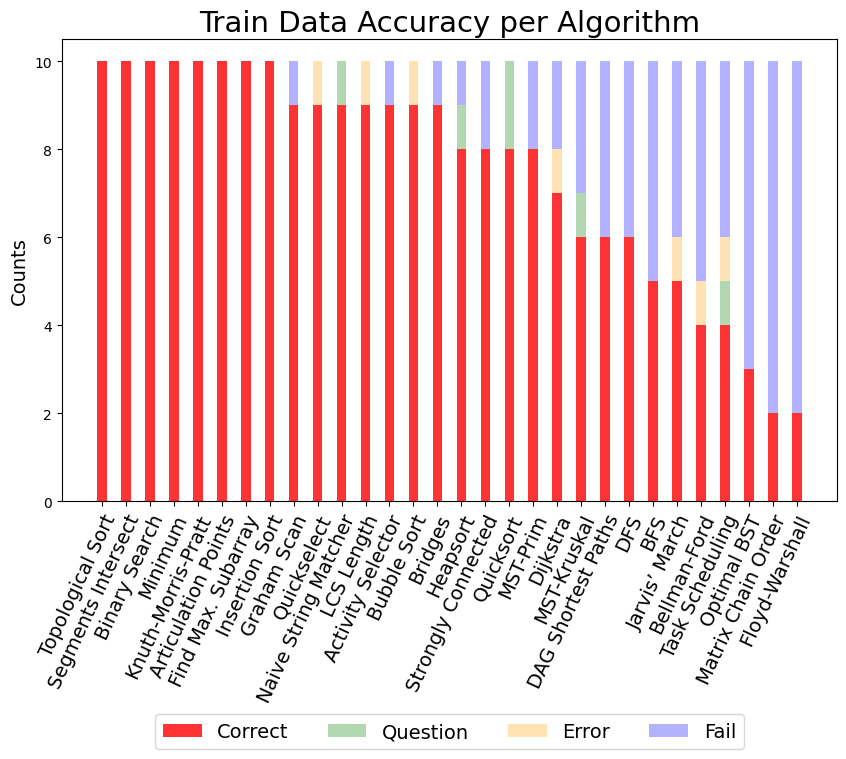}
    \caption{ 
    A breakdown of outcomes organized by problem type. We show the portion of samples that are correct along with three failure types indicated by colour from 10 examples from testing (left) and training (right) splits of the CLRS benchmark \citep{clrs-velickovic}.}
    \label{fig:acc}
\end{figure}

The particular interface we use is the OpenAI Assistant API, specifically the \texttt{gpt-4-preview-1106} model with code interpreter and file upload. 
When arrays are large enough that they do not fit in the context window of the Assistant, we need to provide the data in auxiliary text files.
The Assistant API is in Beta testing for these experiments, so we also count errors caused by the infrastructure, e.g. the Assistant being unable to run Python code or being unable to find an uploaded text file.
See \Cref{fig:acc}
for counts of various types of failure cases.
Anecdotally, we find that asking for output arrays through text files allows for better performance with ChatGPT (and it makes the evaluation process much simpler). 

Importantly, we are not clamining that ChatGPT is incapable of achieving better performance on these tasks if it were given more detail in the prompt or feedback from the user.
Our experiments cover only single prompting with a minimal amount of description required to expect good results. Interestingly, this is already enough to achieve state-of-the-art performance on the benchmark.

\paragraph{Open source models} We also prompt two open source models: Llama-2-70b-chat-hf \citep{touvron2023llama} and Zephyr-7B-beta \citep{tunstall2023zephyr}.
We find that these models perform much worse overall. 
They do not have access to a code interpreter, which is likely the cause of frequent errors over long reasoning trajectories.
However, even on the training problems (smaller problems, fewer nodes in the graph or entries in the list), these models fail to meet the strict notion of success.
They often show peculiar behaviour such as describing and executing the algorithm incorrectly but giving the correct final solution.

\section{Discussion}
What does it mean to learn an algorithm from data? 
Recent works that tackle this question pose various answers.
The GNN community has the CLRS benchmark and focuses on building and training GNNs that perform known algorithms, like breadth-first search, step by step \citep{ibarz2022generalistlearner, mirjanic2023latent} and can extrapolate by executing the same algorithm on larger inputs.

These algorithm problems stand in stark contrast to image classification or protein folding, where neural networks are state of the art.
Neural networks are certainly not an optimal choice for representing algorithms as they have no accuracy guarantees and they come with higher compute requirements than classical algorithms.
This raises a question about why building small specialist models, like neural algorithmic reasoners, is of value.
To this we have two compelling answers:
\begin{enumerate}[label=\Roman*)]
    \item While some of this work focuses on known algorithms for now, the promising direction is to build learning pipelines that are capable of discovering and executing novel algorithmic approaches.
    \item While large multi-modal models currently dominate at many tasks, specialist models take less compute and outperform generalists in some cases.
\end{enumerate}

Algorithm design can range from handcrafted hard coded algorithms, like implementing breadth-first search in Python, all the way to learned approaches from data without any oversight.
On one end of this range sits neural networks that can extract scalable processes from data alone (with no particular algorithm chosen \emph{a priori}).
Training these networks amounts to finding a function that executes a process like search on inputs of arbitrary size.
For example, \citet{bansal2022endtoend} show that recurrent convolutional networks can learn to extract scalable reasoning processes to solve problems like computing prefix sums and solving mazes from inputs and outputs alone.
This line of work suggests that specialist neural networks can help us discover new algorithmic approaches.
In fact, new faster sorting algorithms are already being discovered using deep reinforcement learning \citep{mankowitz2023faster}.

Between handcrafted algorithms and the end-to-end learned algorithms is a large body of work on GNNs where networks are trained to execute specific algorithms
\citep{Velikovi2020Neural, bevilacqua2023neural, rodionov2023neural}.
Rather than training data consisting of inputs and outputs only, like an unsorted list and the corresponding sorted list, these routines train networks to output the intermediate stages of known algorithms, encouraging them to mimic some specific process.
With our results from experiments with ChatGPT, we now have another kind of AI system that can take these CLRS benchmark problems as input and return the correct answer as output, an example of a generalist model that can compete with specialist models. 
However, the small specialist models require orders of magnitude less training than ChatGPT.
Moreover, in some areas there are specialist models that are far ahead of LLMs, like protein folding and AlphaFold~\citep{jumper2021highly}.

\paragraph{The value of out of distribution testing}
In some ways, it is difficult to compare results with web-scale models to models with limited training data.
In particular, recall that the CLRS benchmark suite is designed for \emph{out-of-distribution} testing of GNNs. 
This means it is intended to help researchers further understand the difference between generalization within the training domain and extrapolation beyond that domain (in a very controlled shift).
On the other hand, the algorithms to solve the CLRS problems are standard introductory computer science material and we can confidently assume that descriptions of the algorithms, examples of the problems, and exact implementations are likely in ChatGPT's training data.
ChatGPT's performance, therefore, may not be an out-of-distribution result.
Also, GNN models are trained from scratch on CLRS training data, but a language model approach to solving these problems requires pretraining and finetuning, which means it requires much more data and compute.
Our aim is not to serve as a baseline that small specialist models trained as part of experiments on extrapolation need to beat, but rather to formalize the claim that massive pretrained LLMs have a grip on classical algorithms by using popular and accepted benchmarks in the algorithmic reasoning space.
We are answering widely applicable questions about how well state-of-the-art chatbots with code interpreters can handle academic benchmark problems.

\paragraph{Limitations} 
Interestingly, ChatGPT seems to struggle with dynamic programming algorithms.
We see in Figures \ref{fig:test-comp}, \ref{fig:train-F1} and \ref{fig:acc}, that Matrix Chain Order and Optimal Binary Search Tree, two of the three dynamic algorithms in the benchmark, are among the the problems ChatGPT performs the worst on.
Perhaps there are better approaches to solving these dynamic programming problems and ChatGPT has a hard time executing a less-than-optimal algorithm, as is the case with the Task Scheduling Problem; or there may be better prompting strategies for these dynamic programming tasks.
These impacts may also be amplified as we request the split and root matrices as output (for Matrix Chain Order and Optimal Binary Search Tree, respectively), which constitute contrived ways of evaluating execution of an algorithm.
Specifically, these large matrices may be difficult for ChatGPT to manipulate and comprehend without error.
Finally, some features of the ChatGPT version we use (like code execution) are still in Beta testing, so we often run into unexpected errors when using the code interpreter and file retrieval tools.

\paragraph{Future Work}
We only explore single prompts, but perhaps with follow up prompts can help chatbots do even better on these algorithmic problems.
Also, LLMs provide the option to request explanations of the decisions a model makes, either in follow up prompts or by explicitly asking in the initial prompt.
This behavior is unlike specialist GNN models and encourages future work on the transparency of decisions and failure cases.

Even if executing Python code sounds like cheating on the CLRS benchmark, it is a tool at ChatGPT's disposal and seeing it work for problems like this shows a large gap between ChatGPT and the open source models (Llama-2-70b-chat-hf \citep{touvron2023llama} and Zephyr-7B-beta \citep{tunstall2023zephyr}) we test.
Open source language models are lagging behind proprietary closed source models in terms of tool utilization.
These are rich directions for future work in developing both the generalist models that can use tools and specialist models that help uncover new algorithmic approaches.

\section*{Acknowledgments}
This work was made possible by the ONR MURI program and the AFOSR MURI program. Commercial support was provided by Capital One Bank, the Amazon Research Award program, and Open Philanthropy. Further support was provided by the National Science Foundation (IIS-2212182), and by the NSF TRAILS Institute (2229885).
We thank Petar Veli{\v{c}}kovi{\'c} for helpful and constructive feedback on initial drafts of this work.

\bibliography{colm2024_conference}
\bibliographystyle{colm2024_conference}

\appendix

\footnotesize
\section{Additional Example Prompts}
\label{sec:example-prompts}
\normalsize
Example prompts for training split data from the CLRS benchmark, test split prompts are almost identical but include longer prompts that contain the larger data in the context window.

\begin{tcolorbox}[title=Example Prompt: Sorting - Quick Sort, colback=gray!5, colframe=gray!75, rounded corners]
{\footnotesize \textbf{System Prompt:}\\ You are a helpful assistant for solving and explaining classical coding problems.\\
\textbf{Context:} \\Perform Quick Sort on this list [0.28407, 0.65575, 0.98991, 0.72754, 0.41989, 0.24323, 0.80404, 0.70888, 0.10029, 0.82149, 0.39518, 0.9323, 0.87058, 0.8907, 0.24001, 0.01178] and output the order of the indices, starting with initial indices [0, 1, 2, 3, 4, 5, 6, 7, 8, 9, 10, 11, 12, 13, 14, 15]. I cannot run code. You should show as much work as possible, at least the first step, and run until the sorting process is complete. The last line of your output should be the solution to the problem. If this is from running code, you should restate the output in our conversation.\\\\
\textbf{Output:}\\
To perform a Quick Sort and track the indices, we can follow the standard Quick Sort algorithm while keeping an extra array that holds the indices of the numbers. When we swap elements during the partitioning step, we also swap the corresponding indices.\\
Let's manually go through the first step of partitioning the list using the last element as the pivot and then proceed with the full sorting algorithm.\\
Initial list:
`[0.28407, 0.65575, 0.98991, 0.72754, 0.41989, 0.24323, 0.80404, 0.70888, 0.10029, 0.82149, 0.39518, 0.9323, 0.87058, 0.8907, 0.24001, 0.01178]`\\
Initial indices:
`[0, 1, 2, 3, 4, 5, 6, 7, 8, 9, 10, 11, 12, 13, 14, 15]`\\
The pivot is `0.01178` (last element). We want to partition the array into two parts: elements less than the pivot and elements greater than the pivot. We will keep track of the "less than" section with an index `i`, which will start at -1. For each element in the array (excluding the pivot for now), if it's less than the pivot, we increment `i` and swap the element with the one at `i`.\\
Let's do the first step manually:\\
1. Compare `0.28407` with `0.01178` (pivot): it's greater, so do nothing.
2. Compare `0.65575` with the pivot: it's greater, so do nothing.
3. Continue the same way up to `0.24001`, which is greater than than the pivot as well.\\
At the end of this step, we have not found any elements less than the pivot, so none are swapped and `i` remains at -1. Finally, we swap the pivot with the element at index `i + 1` to put the pivot into its correct sorted position. This will place the pivot at the start of the list.\\
Swapping pivot with the element at index `i + 1` (0th position):
Final list after the first full partition step:
`[0.01178, 0.65575, 0.98991, 0.72754, 0.41989, 0.24323, 0.80404, 0.70888, 0.10029, 0.82149, 0.39518, 0.9323, 0.87058, 0.8907, 0.24001, 0.28407]`\\
Corresponding indices:
`[15, 1, 2, 3, 4, 5, 6, 7, 8, 9, 10, 11, 12, 13, 14, 0]`\\
Now that I've demonstrated the manual process for the first partitioning step, I will implement the full Quick Sort algorithm with index tracking in Python and execute it to find the sorted order of the indices for the entire list. Let's proceed with that.\\
The Quick Sort algorithm has sorted the list, and the corresponding order of the indices starting with initial indices `[0, 1, 2, 3, 4, 5, 6, 7, 8, 9, 10, 11, 12, 13, 14, 15]` is now:
`[15, 8, 14, 5, 0, 10, 4, 1, 7, 3, 6, 9, 12, 13, 11, 2]`\\
This is the final sorted order of the indices, corresponding to the sorted list of numbers.
}
\end{tcolorbox}

\begin{tcolorbox}[title=Example Prompt: Searching - Binary Search, colback=gray!5, colframe=gray!75, rounded corners]
{\footnotesize \textbf{System Prompt:}\\ You are a helpful assistant for solving and explaining classical coding problems.\\
\textbf{Context:} \\Perform Binary Search on this list [0.09565, 0.13575, 0.28485, 0.29262, 0.34421, 0.35567, 0.38662, 0.446, 0.46018, 0.51342, 0.51639, 0.55465, 0.66388, 0.66797, 0.71429, 0.7656] to find the first element greater than or equal to 0.07657890766859055 and output the index of this element, starting with initial indices [0, 1, 2, 3, 4, 5, 6, 7, 8, 9, 10, 11, 12, 13, 14, 15]. I cannot run code. You should show as much work as possible, at least the first step, and run until the process is complete. The last line of your output should be the solution to the problem. If this is from running code, you should restate the output in our conversation.
}
\end{tcolorbox}

\begin{tcolorbox}[title=Example Prompt: Greedy - Activity Selector, colback=gray!5, colframe=gray!75, rounded corners]
{\footnotesize \textbf{System Prompt:}\\ You are a helpful assistant for solving and explaining classical coding problems.\\
\textbf{Context:} \\
Solve the Activity Selection problem, of selecting a maximum-size set of mutually compatible activities, of these activities. With start times [0.33228, 0.00666, 0.11203, 0.07502, 0.04902, 0.20208, 0.62735, 0.42529, 0.41026, 0.46971, 0.69535, 0.30077, 0.59919, 0.66007, 0.13262, 0.00406] and finish times [0.47054, 0.78444, 0.98931, 0.8614, 0.88159, 0.36654, 0.93642, 0.79705, 0.46615, 0.54918, 0.76103, 0.98386, 0.94017, 0.73399, 0.24703, 0.64587] output the indices of the selected activities, starting with initial indices [0, 1, 2, 3, 4, 5, 6, 7, 8, 9, 10, 11, 12, 13, 14, 15]. You should sort the list of selected activities in ascending order before returning them. I cannot run code. You should show as much work as possible, at least the first step, and run until the process is complete. The last line of your output should be the solution to the problem. If this is from running code, you should restate the output in our conversation.
}
\end{tcolorbox}

\begin{tcolorbox}[title=Example Prompt: Divide and Conquer - Maximum Subarray, colback=gray!5, colframe=gray!75, rounded corners]
{\footnotesize \textbf{System Prompt:}\\ You are a helpful assistant for solving and explaining classical coding problems.\\
\textbf{Context:} \\Find the Maximum Subarray of this list [0.25385, 0.28883, -0.78755, -0.40205, -0.22093, 0.90873, 0.15351, -0.82528, -0.78568, 0.90737, 0.61069, -0.1434, 0.34498, -0.57614, 0.04033, 0.95108] output the start and end index of the subarray, starting with initial indices [0, 1, 2, 3, 4, 5, 6, 7, 8, 9, 10, 11, 12, 13, 14, 15]. I cannot run code. You should show as much work as possible, at least the first step, and run until the process is complete. The last line of your output should be the solution to the problem. If this is from running code, you should restate the output in our conversation.
}
\end{tcolorbox}

\begin{tcolorbox}[title=Example Prompt: Graph Problems - Strongly Connected Components, colback=gray!5, colframe=gray!75, rounded corners]
{\footnotesize \textbf{System Prompt:}\\ You are a helpful assistant for solving and explaining classical coding problems.\\
\textbf{Context:} \\
Find the Strongly Connected Components of this adjacency matrix [[0, 0, 0, 0, 0, 0, 0, 0, 0, 0, 0, 0, 1.0, 0, 0, 0], [0, 1.0, 0, 0, 0, 0, 1.0, 0, 0, 0, 0, 0, 0, 0, 0, 0], [0, 0, 0, 0, 0, 0, 0, 0, 0, 0, 0, 0, 0, 1.0, 0, 1.0], [0, 0, 0, 1.0, 0, 0, 0, 1.0, 0, 0, 0, 0, 0, 0, 0, 0], [0, 0, 0, 0, 1.0, 0, 0, 0, 1.0, 1.0, 0, 0, 0, 0, 0, 0], [0, 0, 0, 0, 0, 1.0, 0, 0, 0, 0, 1.0, 0, 1.0, 1.0, 0, 0], [0, 1.0, 0, 0, 0, 0, 0, 0, 0, 0, 0, 0, 0, 0, 0, 0], [0, 1.0, 0, 1.0, 0, 0, 0, 1.0, 0, 0, 0, 0, 0, 0, 0, 0], [0, 0, 0, 0, 0, 0, 0, 0, 0, 0, 0, 1.0, 0, 0, 0, 0], [0, 0, 0, 0, 1.0, 0, 0, 0, 1.0, 1.0, 0, 1.0, 0, 0, 0, 0], [0, 0, 0, 0, 0, 1.0, 0, 0, 0, 0, 0, 0, 0, 0, 0, 0], [0, 0, 0, 0, 1.0, 0, 0, 0, 1.0, 0, 0, 0, 0, 0, 0, 0], [1.0, 0, 0, 0, 0, 1.0, 0, 0, 0, 0, 1.0, 0, 1.0, 0, 0, 0], [0, 0, 1.0, 0, 0, 0, 0, 0, 0, 0, 0, 0, 0, 1.0, 1.0, 0], [0, 0, 0, 0, 0, 0, 0, 0, 0, 0, 0, 0, 0, 1.0, 0, 0], [0, 0, 1.0, 0, 0, 0, 0, 0, 0, 0, 0, 0, 0, 1.0, 0, 1.0]], using Depth First Search, return a list of lists where each sublist contains the indices of the nodes in that strongly connected component. If there is a choice between nodes, always pick the one with lower value. I cannot run code. You should show as much work as possible, at least the first step, and run until the process is complete. The last line of your output should be the solution to the problem. If this is from running code, you should restate the output in our conversation.
}
\end{tcolorbox}

\begin{tcolorbox}[title=Example Prompt: String Problems - Naive String Matcher, colback=gray!5, colframe=gray!75, rounded corners]
{\footnotesize \textbf{System Prompt:}\\ You are a helpful assistant for solving and explaining classical coding problems.\\
\textbf{Context:} 
Perform the naive string matching algorithm on [[0, 1, 0, 0], [0, 0, 1, 0], [1, 0, 0, 0], [0, 0, 1, 0], [0, 1, 0, 0], [0, 0, 1, 0], [0, 0, 0, 1], [1, 0, 0, 0], [0, 1, 0, 0], [1, 0, 0, 0], [1, 0, 0, 0], [0, 1, 0, 0], [0, 1, 0, 0], [0, 0, 0, 1], [1, 0, 0, 0], [0, 0, 0, 1]] and [[0, 0, 1, 0], [0, 1, 0, 0], [0, 0, 1, 0], [0, 0, 0, 1]], where the characters of the string are one hot encoded from a size 4 vocabulary. Return the beginning index at which the strings overlap. If you write python code, the first code block should only be you defining the strings. I cannot run code. You should show as much work as possible, at least the first step, and run until the process is complete. The last line of your output should be the solution to the problem. If this is from running code, you should restate the output in our conversation.
}
\end{tcolorbox}

\begin{tcolorbox}[title=Example Prompt: Geometry problems - Jarvis' March, colback=gray!5, colframe=gray!75, rounded corners]
{\footnotesize \textbf{System Prompt:}\\ You are a helpful assistant for solving and explaining classical coding problems.\\
\textbf{Context:} \\
Perform the Jarvis March Algorithm on these points, X coordinates [1.2194, -1.11406, 0.38929, -1.73849, -0.31843, 1.22709, 0.43665, 0.7779, -1.62778, -0.26118, -0.24323, -0.66371, 0.81454, -1.17166, -0.03785, 1.07014], Y coordinates [1.498, -1.25286, 0.34116, 0.53362, -0.23869, 0.35766, -1.86391, 0.53266, -0.29587, 1.28856, -1.34246, -1.10064, 1.74479, -0.59935, 0.48395, 1.55081], return the indices of the points in the hull, sorting these indices in ascending order when printing, indexing from 0. If you write python code, the first code block should only be you defining the two arrays. I cannot run code. You should show as much work as possible, at least the first step, and run until the process is complete. The last line of your output should be the solution to the problem. If this is from running code, you should restate the output in our conversation.
}
\end{tcolorbox}

\subsection{Dynamic Programming - Longest Common Subsequence}
\label{subsec:lcs}
\normalsize
For Longest Common Subsequence, we accept multiple outputs. We mark ChatGPT correct when it generates either of the examples in the Example Ground Truth box below.
Note that this is the only problem where we stray from the benchmarks output format, since ChatGPT is capable of generating arrow characters and we find that it in fact uses this style of output more often.
\begin{tcolorbox}[title=Example Prompt: Dynamic Programming - Longest Common Subsequence, colback=gray!5, colframe=gray!75, rounded corners]
{\footnotesize \textbf{System Prompt:}\\ You are a helpful assistant for solving and explaining classical coding problems.\\
\textbf{Context:} \\
Find the longest common subsequence between [[1, 0, 0, 0], [1, 0, 0, 0], [0, 1, 0, 0], [0, 1, 0, 0], [0, 1, 0, 0], [0, 0, 0, 1], [0, 1, 0, 0], [1, 0, 0, 0]], and string [[0, 0, 1, 0], [0, 0, 0, 1], [0, 0, 0, 1], [0, 1, 0, 0], [0, 0, 0, 1], [1, 0, 0, 0], [0, 0, 0, 1], [0, 0, 0, 1]], where the characters of the string are one hot encoded from a size 4 vocabulary. Return the full b matrix from the dynamic program using these characters \(\nwarrow\), ↑, ←, as a txt file. If there is a choice between ↑ and ←, choose ↑. If you write python code, the first code block should only be you defining the strings. I cannot run code. You should show as much work as possible, at least the first step, and run until the process is complete. The last line of your output should be the solution to the problem. If this is from running code, you should restate the output in our conversation.
}
\end{tcolorbox}
\begin{tcolorbox}[title=Ground Truth Answers, colback=gray!5, colframe=gray!75, rounded corners]
{\footnotesize
\textbf{Ground Truth:} \\
$$
\begin{bmatrix}
  1. & 1. & 1. & 1. & 1. & 3. & 2. & 2.\\
  1. & 1. & 1. & 1. & 1. & 3. & 1. & 1.\\
  1. & 1. & 1. & 3. & 2. & 1. & 1. & 1.\\
  1. & 1. & 1. & 3. & 1. & 1. & 1. & 1.\\
  1. & 1. & 1. & 3. & 1. & 1. & 1. & 1.\\
  1. & 3. & 3. & 1. & 3. & 2. & 3. & 3.\\
  1. & 1. & 1. & 3. & 1. & 1. & 1. & 1.\\
  1. & 1. & 1. & 1. & 1. & 3. & 2. & 2.\\
\end{bmatrix}
\begin{bmatrix}
  \uparrow & \uparrow & \uparrow & \uparrow & \uparrow & \nwarrow & \leftarrow & \leftarrow\\
  \uparrow & \uparrow & \uparrow & \uparrow & \uparrow & \nwarrow & \uparrow & \uparrow\\
  \uparrow & \uparrow & \uparrow & \nwarrow & \leftarrow & \uparrow & \uparrow & \uparrow\\
  \uparrow & \uparrow & \uparrow & \nwarrow & \uparrow & \uparrow & \uparrow & \uparrow\\
  \uparrow & \uparrow & \uparrow & \nwarrow & \uparrow & \uparrow & \uparrow & \uparrow\\
  \uparrow & \nwarrow & \nwarrow & \uparrow & \nwarrow & \leftarrow & \nwarrow & \nwarrow\\
  \uparrow & \uparrow & \uparrow & \nwarrow & \uparrow & \uparrow & \uparrow & \uparrow\\
  \uparrow & \uparrow & \uparrow & \uparrow & \uparrow & \nwarrow & \leftarrow & \leftarrow\\
\end{bmatrix}
$$
}
\end{tcolorbox}
\normalsize
These alternatives arise as a result of the original constraint for GNNs: that the inputs and outputs are one-hot encodings.
The output is originally a large three-dimensional array representing one-hot encodings at each location in the matrix. 
In our CLRS4LM codebase we reduce this to a two-dimensional array with integers at each location instead of one-hot encodings.
Moreover, as language models can output unicode characters we also include the output matrix shown in the CLRS textbook and taught in most algorithms classes which uses arrows ($\leftarrow, \;\nwarrow,\; \uparrow$) instead of of integer indices.
\begin{figure}[ht!]
    \centering
    \includegraphics[trim=0cm 10cm 0cm 10cm,width=0.7\textwidth,clip]{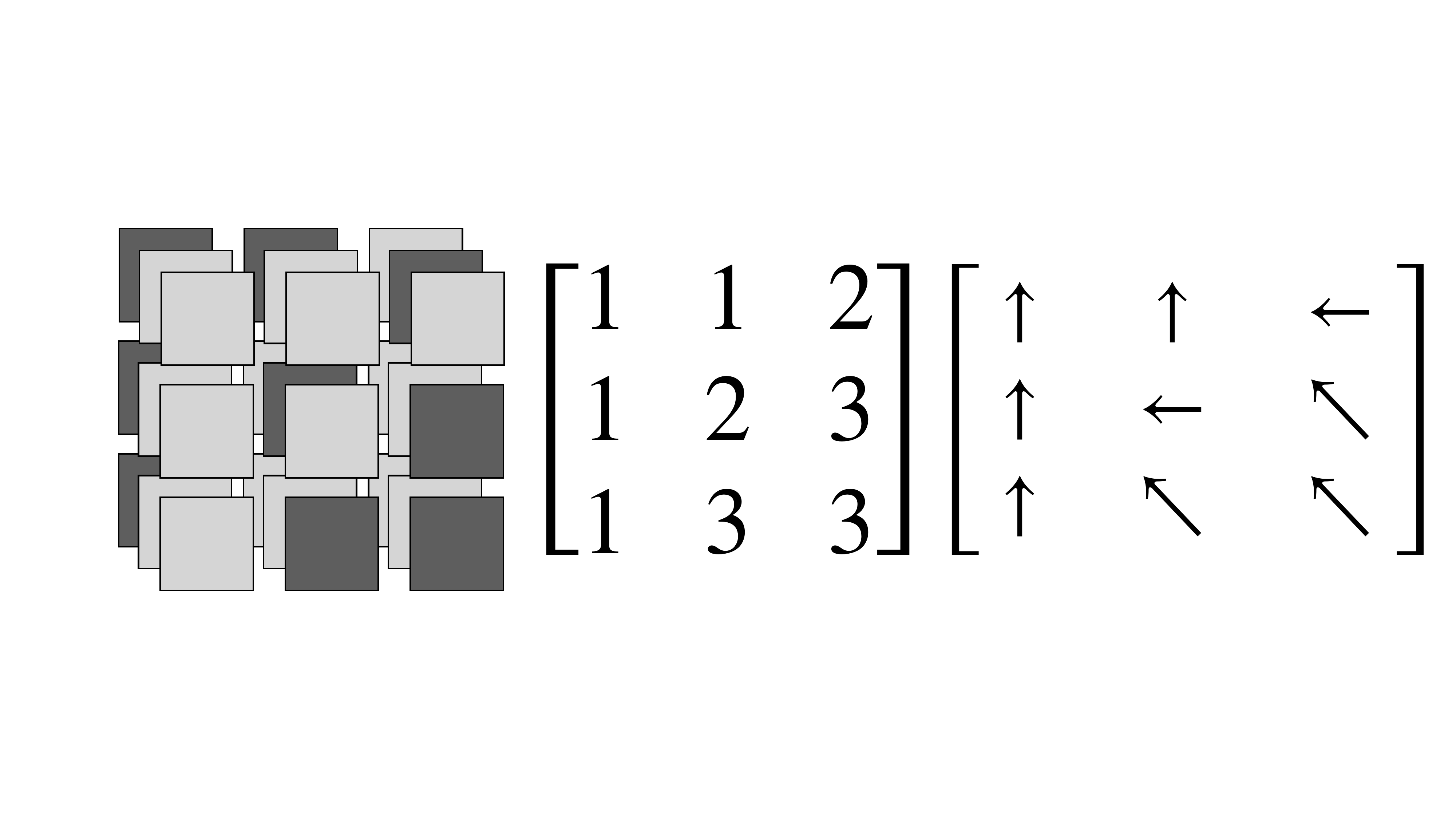}
    \caption{\textbf{Left:} A graphical representation of one-hot encodings at every location in a $3\times3$ matrix. \textbf{Center:} The corresponding matrix of integer-valued indices. \textbf{Right:} The unicode style arrow-based representation of the same matrix.}
    \label{fig:sub-seq}
\end{figure}


\end{document}